\documentclass{article}

\usepackage{microtype}
\usepackage{graphicx}
\usepackage{subcaption}
\usepackage{booktabs} % for professional tables
\usepackage{multirow} 
\usepackage[table]{xcolor}
\usepackage{multicol}
\usepackage{amsmath}
\allowdisplaybreaks

\usepackage{hyperref}

\DeclareMathAlphabet{\mathbbold}{U}{bbold}{m}{n}

\usepackage[preprint]{icml2026}

\usepackage{amsmath}
\usepackage{amssymb}
\usepackage{mathtools}
\usepackage{amsthm}

\usepackage[capitalize,noabbrev]{cleveref}

\theoremstyle{plain}
\newtheorem{theorem}{Theorem}[section]

\theoremstyle{definition}
\newtheorem{definition}[theorem]{Definition}

\theoremstyle{remark}

\usepackage[textsize=tiny]{todonotes}

\DeclareMathOperator*{\argmin}{arg\,min}

\icmltitlerunning{Submission and Formatting Instructions for ICML 2026}

\begin{document}

\twocolumn[
  \icmltitle{GaussianPSL: Soft partitioning for complex PSL problem}

  % It is OKAY to include author information, even for blind submissions: the
  % style file will automatically remove it for you unless you've provided
  % the [accepted] option to the icml2026 package.

  % List of affiliations: The first argument should be a (short) identifier you
  % will use later to specify author affiliations Academic affiliations
  % should list Department, University, City, Region, Country Industry
  % affiliations should list Company, City, Region, Country

  % You can specify symbols, otherwise they are numbered in order. Ideally, you
  % should not use this facility. Affiliations will be numbered in order of
  % appearance and this is the preferred way.
  \icmlsetsymbol{equal}{*}

  \begin{icmlauthorlist}
    \icmlauthor{Mai-Phuong Dinh}{yyy}
    \icmlauthor{Van-Nam Huynh}{yyy}
    
  \end{icmlauthorlist}

  \icmlaffiliation{yyy}{School of Information Science, Japan Advanced Institute of Science and Technology}

  \icmlcorrespondingauthor{Van-Nam Huynh}{huynh@jaist.ac.jp}

  % You may provide any keywords that you find helpful for describing your
  % paper; these are used to populate the "keywords" metadata in the PDF but
  % will not be shown in the document
  \icmlkeywords{Machine Learning, ICML}

  \vskip 0.3in
]

% this must go after the closing bracket ] following \twocolumn[ ...

% This command actually creates the footnote in the first column listing the
% affiliations and the copyright notice. The command takes one argument, which
% is text to display at the start of the footnote. The \icmlEqualContribution
% command is standard text for equal contribution. Remove it (just {}) if you
% do not need this facility.

% Use ONE of the following lines. DO NOT remove the command.
% If you have no special notice, KEEP empty braces:
\printAffiliationsAndNotice{}  % no special notice (required even if empty)
% Or, if applicable, use the standard equal contribution text:
% \printAffiliationsAndNotice{\icmlEqualContribution}

\begin{abstract}
Many practical applications of multi-objective optimization (MOO), including engineering design, autonomous systems, and machine learning, often yield complex Pareto frontiers (e.g., discontinuous, degenerate, or non-convex), which pose challenges for traditional scalarization and Pareto Set Learning (PSL) methods that struggle to approximate them accurately. In this paper, we propose GaussianPSL, a novel framework that uses soft partitions of the Pareto decision/objective space to address the challenges posed by complex Pareto frontiers. Our method dynamically partitions the space, enabling simple MLP networks to learn localized features within each region and then aggregate this information for the final prediction. This partition-aware strategy enhances both exploration and convergence, reduces sensitivity to initialization, and improves robustness against local optima. Experimental results demonstrate that the proposed approach consistently outperforms standard PSL models in learning complex Pareto fronts while maintaining model simplicity. Overall, GaussianPSL offers a new direction for effective, scalable MOO in challenging frontier geometries.
\end{abstract}

\section{Introduction}
Multi-objective optimization (MOO) plays a critical role in addressing complex decision-making problems where multiple conflicting objectives must be optimized simultaneously. Many real-world applications, from engineering design~\cite{Sinha2018, LIU2025119541} and control systems to machine learning~\cite{Rai01012003} and autonomous decision-making~\cite{pmlr-v119-mahapatra20a, SU2024101605}, involve complex Pareto frontiers, where optimal trade-offs form intricate, non-convex surfaces~\cite{Miettinen1999}. In such settings, finding diverse and representative solutions becomes especially challenging, as traditional methods (e.g, Evolutionary Multi-Objective Optimization~\cite{zheng2025weakparetoboundaryachilles}) often fail to capture the full spectrum of Pareto-optimal solutions, particularly in non-convex regions. These challenges underscore the need for advanced MOO methods that can explore non-convex solution spaces~\cite{huang2025optimizationweaklyparetoset}, maintain diversity across the Pareto front, and produce solutions aligned with real-world constraints and preferences. Some work has been done, for example, Ye et al.~\cite{Ye2024EvolutionaryPS} leverage genetic algorithms (GAs) for preference vector selection to improve neural network performance by focusing on salient regions of the frontier while reducing the impact of weak-order noise. But it is not fully efficient and does not address the core problem of limitations caused by weak Pareto regions in gradient-based models. 

\textbf{Challenge.} Pareto Set Learning (PSL) models typically perform well on convex Pareto frontiers, as reported in the previous studies~\cite{ pmlr-v119-ma20a, zhang2023, LI2023101253, chen2025}. However, rarely does any work explicitly aim to target complex Pareto frontiers such as non-convex, degenerate, and discontinuous fronts, which are common in real-world problems~\cite{Tanabe_2020}. 

\textbf{Approach.} In this paper, we present a novel Pareto Set Learning framework called GaussianPSL that incorporates soft partitioning in the optimal decision space. We apply the Adaptive Density Control techniques~\cite{kerbl3Dgaussians} to dynamically manage these partitions. While the MLP network learns the general information of the space, layers of Logistic Regression try to learn the local information of each partition. This approach results in a more robust Pareto set, enhancing exploration while ensuring convergence to the optimal front. Our approach tackles the challenge of local optima created by weak Pareto optimal areas, where optimization performance is often highly sensitive to initialization. Our key contributions are as follows:
\begin{itemize}
\item We first revisit the mathematical formulation for learning a controllable Pareto set in the context of multi-objective optimization, and then reveal the limitations of the PSL model when handling complex frontiers with a simple example.
\item We then introduce GaussianPSL, a novel framework for Pareto Set Learning that incorporates a soft partitioning mechanism to better handle complex Pareto frontiers.
\item Finally, we conduct extensive experiments on both synthetic and real-world multi-objective optimization problems to evaluate effectiveness and robustness (e.g., against the locally Pareto optimal created by weak Pareto optimal regions). 
\end{itemize}

\section{Preliminaries}
In this section, we provide the necessary background on multi-objective optimization, Pareto optimality, and the standard formulation of the vanilla Pareto Set Learning model. 
\subsection{Multi-Objective Optimization}
Given the feasible set $\mathbf{X} \subseteq \mathbb{R}^n$ of decision vectors $\mathbf{x}$, the vector-valued objective function is defined as:
\begin{equation} \label{eq:moo_def}
	\begin{split}
		f : \quad & \mathbf{X} \rightarrow \mathbb{R}^m \\
		   & \mathbf{x} \mapsto f(\mathbf{x}) = \left( f_1(\mathbf{x}),  f_2(\mathbf{x}), \cdots, f_m(\mathbf{x}) \right)
	\end{split}
\end{equation}  
Without loss of generality, assuming all objective functions in Equation (\ref{eq:moo_def}) are to be minimized, then the multi-objective optimization problem can be formulated as:
\begin{equation}  \label{eq:moo_opt}
	 \min_{\mathbf{x} \in \mathbf{X}} f(\mathbf{x}) = \min_{\mathbf{x} \in \mathbf{X}} (f_{1}(\mathbf{x}),f_{2}(\mathbf{x}),\ldots ,f_{m}(\mathbf{x}))
\end{equation} 
whereas $\mathbf{X}$ is a feasible solution space, and the set $\mathbf{Y} = \{ \mathbf{y} = f(\mathbf{x}) \mid \mathbf{x} \in \mathbf{X} \}$ represents the objective space, i.e., the image of $\mathbf{X}$ under the mapping $f$. Each $\mathbf{y} = f(\mathbf{x})$ is referred to as an objective vector or outcome of the decision vector. Our problem is to identify a feasible solution $\mathbf{x}$ that minimizes all the objectives simultaneously.   
% [TODO] maximize and minimize checking
\subsection{Pareto Optimality} 
In general, there is no feasible solution that minimizes all objective functions simultaneously. Therefore, attention is directed toward Pareto optimal solutions~\cite{Pareto:1971}, which are solutions that cannot be improved in any objective without degrading at least one of the others.

\begin{definition}[Pareto Dominance]
	An objective vector $\mathbf{y}^1 \in \mathbf{Y}$ is said to dominate another vector $\mathbf{y}^2 \in \mathbf{Y}$, denoted $ \mathbf{y}^1 \succ \mathbf{y}^2$, if and only if:
    
	\begin{itemize}
		\item $\forall \; i \in \{1 \dots m\}, \mathbf{y}^1_i \leq \mathbf{y}^2_i $ and
		\item $\exists \; j \in \{1 \dots m\}, \mathbf{y}^1_j < \mathbf{y}^2_j $
	\end{itemize}
This means that $\mathbf{y}^1$ is at least as good as $\mathbf{y}^2$ in all objectives and strictly better in at least one. The relation $\succ$ is called the Pareto order and defines a strict partial order over $\mathbf{Y}$. 

If $\mathbf{x}^1, \mathbf{x}^2 \in \mathbf{X}$ are feasible solutions such that $f(\mathbf{x}^1) = \mathbf{y}^1$ and $f(\mathbf{x}^2) = \mathbf{y}^2$, then we say $\mathbf{x}^1$ Pareto dominates $\mathbf{x}^2$ and denote this as $\mathbf{x}^1 \succ_f \mathbf{x}^2$.
\end{definition}

\paragraph{Strong Pareto Optimality}
A decision vector $\mathbf{x}^\ast$ is considerated as a \emph{strongly Pareto optimal} if there does not exist any $\mathbf{x} \in \mathbf{X}$ such that $\mathbf{x} \succ_f \mathbf{x}^\ast$
\paragraph{Weak Pareto Optimality.}
A point $\mathbf{x}^\ast \in \mathbf{X}$ is \emph{weakly Pareto optimal} if there does not exist any $\mathbf{x} \in \mathbf{X}$ such that
\[
f_i(\mathbf{x}) < f_i(\mathbf{x}^\ast) \quad \text{for all } i = 1,\dots,m.
\]

\begin{figure}[t]
    \centering
    \begin{subfigure}[b]{0.48\linewidth}
        \centering
        \includegraphics[width=\linewidth]{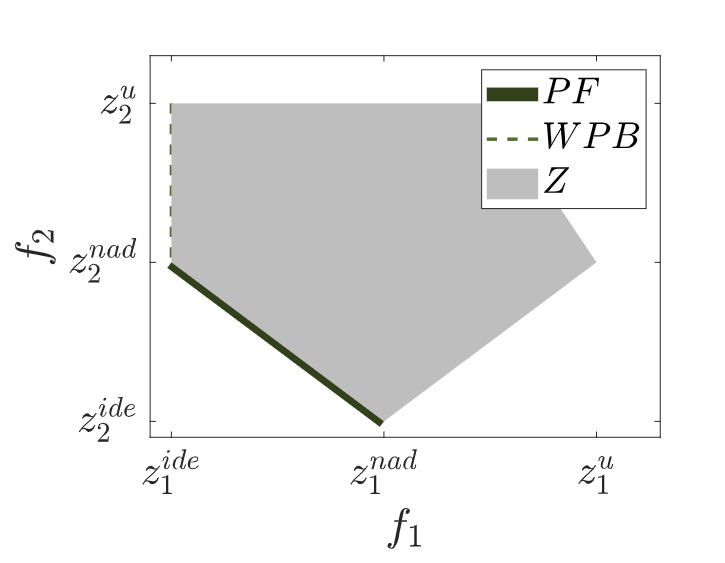}
        \caption{Strong Pareto optimality}
        \label{fig:strong_pareto}
    \end{subfigure}
    \hfill
    \begin{subfigure}[b]{0.45\linewidth}
        \centering
        \includegraphics[width=\linewidth]{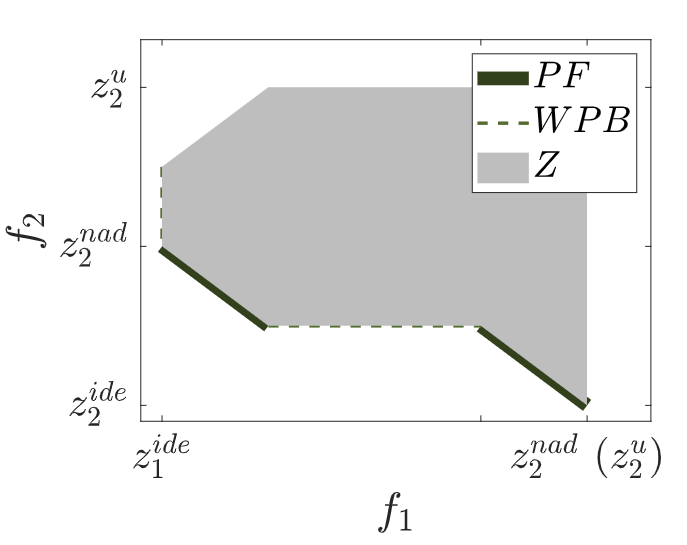}
        \caption{Weak Pareto optimality}
        \label{fig:weak_pareto}
    \end{subfigure}
    \caption{Illustration of strong and weak Pareto optimality for a two-objective minimization problem. Source: \cite{zheng2025weakparetoboundaryachilles}}
    \label{fig:pareto_comparison}
\end{figure}

\begin{definition}[Pareto Frontier and Optimal Solution Space]
	An objective vector $\mathbf{y}^* \in \mathbf{Y}^* \subset \mathbf{Y}$ is said to lie on the Pareto frontier if and only if
	$\nexists \; \mathbf{y}\in \mathbf{Y}$ such that $\mathbf{y} \succ \mathbf{y}^*$. The corresponding set of feasible solutions $\mathbf{X}^* = \{ \mathbf{x} \in \mathbf{X} \mid f(\mathbf{x}) \in \mathbf{Y}^* \}$ is called the Pareto optimal solution space.
\end{definition}

\subsection{Pareto Set Learning}
Pareto Set Learning (PSL) refers to a class of methods that employ neural networks to approximate a mapping from preference vectors $\lambda \in \mathbb{R}^k$ (Appendix \ref{appendix:preference_vector}) to corresponding Pareto-optimal solutions $\mathbf{X}^*$, assuming that the objective function $f$ is known.

In practice, PSL models are typically implemented using neural networks and aim to approximate the Pareto-optimal solution set by minimizing a loss function defined in the objective space. Because this loss is formulated in terms of objective values rather than directly over the decision variables, it acts as a surrogate loss. This indirection introduces additional complexity, rendering the optimization process challenging and non-trivial. A Pareto set model can be formally defined as follows:
\begin{equation} \label{eq:psl_def}
	\begin{split}
		\mathbf{x}_\lambda &= \phi_{\mathbf{\theta}} (\mathbf{\lambda}) \\
		\mathbf{y}_\lambda &= f(\mathbf{x}_\lambda)
	\end{split}
\end{equation}
where ${\lambda} \in \mathbb{R}^m$ is the preference vector, $\phi_{\mathbf{\theta}}: \mathbb{R}^m \rightarrow \mathbb{R}^n$ is the Pareto set model, parameterized by $\mathbf{\theta}$, $f: \mathbb{R}^n \rightarrow \mathbb{R}^m$ is the vector-valued objective function, $\mathbf{x}_\lambda$ is the predicted decision vector, and $\mathbf{y}_\lambda$ is the corresponding objective vector. 

During training, the model learns to map each preference vector $\mathbf{\lambda}$ to a decision vector $\mathbf{x}_\lambda$ such that the resulting objective vector $f(\mathbf{x}_\lambda)$ lies on or near the Pareto frontier. To achieve this, a scalarization function $s(\cdot)$ is applied to the objective values, converting the multi-objective problem into a single-objective one. The optimization goal of PSL is to minimize the expected scalarized objective over a predefined distribution $\mathcal{D}$ of preference vectors:
\begin{equation} \label{eq:psl_opt}
	\begin{split}
		\theta^* &= \argmin_\mathbf{\theta} \mathbb{E}_{\mathbf{\lambda} \sim \mathcal{D}} [s(f(\phi_\mathbf{\theta} (\mathbf{\lambda}))] \\ &= \argmin_\mathbf{\theta} \mathbb{E}_{\mathbf{\lambda} \sim \mathcal{D}} [s(f(\mathbf{x}_\mathbf{\lambda}))]
	\end{split}	
\end{equation}

\section{Motivation}
We begin by examining the shortcomings of PSL on weak Pareto frontiers and demonstrate these issues using a representative case of a disconnected Pareto front.
\subsection{Limitations of PSL on the weak Pareto frontier}
For differentiable objectives, define the gradient matrix
\[
\nabla f(\mathbf{x}) := \big[\nabla f_{1}(\mathbf{x}),\dots,\nabla f_{m}(\mathbf{x})\big] \in \mathbb{R}^{n\times m}.
\]

A standard first-order notion is \emph{Pareto stationarity} \cite{Ye2021ParetoNG}:
\begin{equation}
\begin{gathered}
    \min_{\lambda\in\Delta_{m}} \big\|\nabla f(\mathbf{x})\lambda\big\| = 0,\\
    \Delta_{m} := \{\lambda\in\mathbb{R}^{m}_{+} : \textstyle\sum_{i=1}^{m}\lambda_{i}=1\}
\end{gathered}
\label{eq:pareto_stationary}   
\end{equation}

This means there exists \(\lambda^{*}\in\Delta_{m}\) with
\begin{equation}
    \nabla f(\mathbf{x})\lambda^{*} = \sum_{i=1}^{m}\lambda^{*}_{i}\,\nabla f_{i}(\mathbf{x}) = 0.
\label{eq:pareto_stationary_2}
\end{equation}

Equivalently, there is \emph{no common descent direction}:
\begin{equation}
    \nexists d\in\mathbb{R}^{n}:\ \nabla f_{i}(\mathbf{x})^{\top} d < 0\ \forall i.
\label{eq:pareto_stationary_3}
\end{equation}

Under standard convexity assumptions on \(f_{i}\), conditions (\ref{eq:pareto_stationary})-(\ref{eq:pareto_stationary_3}) are necessary and sufficient for \(\mathbf{x}\) to be \emph{weakly} Pareto optimal. Thus, under convexity:
\[
\text{Pareto stationarity} \Longleftrightarrow \text{weak Pareto optimality}
\]

So any gradient method whose stopping rule is “(\ref{eq:pareto_stationary}) holds” converges (at best) to weak Pareto points.

For Pareto set learning, we parameterize decisions as \(\mathbf{x} = \phi_{\theta}(\lambda)\) and train \(\phi\) (and possibly \(\lambda\)) using gradients so that the set \(\{\phi_{\theta}(\lambda)\}\) approximates the Pareto set.

Because training uses first-order information, a typical inner update is some variant of
\begin{equation}
    \mathbf{x}_{t+1} = \mathbf{x}_{t} - \eta \sum_{i=1}^{m}\lambda^{(t)}_{i}\,\nabla f_{i}(\mathbf{x}_{t}),
    \label{eq:pareto_stationary_5}
\end{equation}

with \(\lambda^{(t)}\in\Delta_{m}\). When updates stop, we have approximately
\begin{equation}
\sum_{i=1}^{m}\lambda^{*}_{i}\,\nabla f_{i}(\mathbf{x}^{*}) \approx 0,\quad \lambda^{*}\in\Delta_{m},
\label{eq:pareto_stationary_6}
\end{equation}
so \(\mathbf{x}^{*}\) satisfies the Pareto stationarity condition (\ref{eq:pareto_stationary})-(\ref{eq:pareto_stationary_2}), hence is (at best) weakly Pareto.

The problem is that weak Pareto points may still be dominated. There can exist \(\mathbf{x}'\) with
\begin{equation}
f_{i}(\mathbf{x}') \le f_{i}(\mathbf{x}^{*})\ \forall i,\quad \exists j: f_{j}(\mathbf{x}') < f_{j}(\mathbf{x}^{*}),
\label{eq:pareto_stationary_7}
\end{equation}
while (\ref{eq:pareto_stationary_3}) still holds at \(\mathbf{x}^{*}\). Intuitively, the dominating point \(\mathbf{x}'\) may be far in parameter space, so its existence is not visible from local first-order information at \(\mathbf{x}^{*}\); gradients at \(\mathbf{x}^{*}\) only tell you that “no single direction improves all objectives at once”, not that “no better point exists anywhere”.

In a finite set of learned decisions \(\{\mathbf{x}^{(k)}\}_{k=1}^{K}\), the model accepts each \(\mathbf{x}^{(k)}\) as “on the Pareto set” once it satisfies (\ref{eq:pareto_stationary_6}). But (\ref{eq:pareto_stationary_7}) can still hold between two learned decisions, i.e., some \(\mathbf{x}^{(k)}\) is dominated by another \(\mathbf{x}^{(\ell)}\), while both are gradient-stationary in the sense of (\ref{eq:pareto_stationary})-(\ref{eq:pareto_stationary_3}). Formally, you can have:
\[
\begin{aligned}
&\sum_{i}\lambda^{(k)}_{i}\nabla f_{i}\big(\mathbf{x}^{(k)}\big)=0,\quad
\sum_{i}\lambda^{(\ell)}_{i}\nabla f_{i}\big(\mathbf{x}^{(\ell)}\big)=0,\\
&f_{i}\big(\mathbf{x}^{(\ell)}\big)\le f_{i}\big(v^{(k)}\big)\ \forall i,\quad
\exists j: f_{j}\big(\mathbf{x}^{(\ell)}\big) < f_{j}\big(\mathbf{x}^{(k)}\big),
\end{aligned}
\]
so both points satisfy the weak, gradient-based condition, but \(\mathbf{x}^{(k)}\) is strictly worse in at least one objective and no better in any. Because the learning process only checks (\ref{eq:pareto_stationary})-(\ref{eq:pareto_stationary_3}) locally and has only finitely number of decisions, it cannot exclude such dominated but weakly Pareto-stationary points from the learned set, which leads to pollution of the learned Pareto set with weak Pareto points and prevents further exploration in decision space.

\subsection{Example: Disconnected Pareto Front in 2D}
In this section, we illustrate how a gradient-based Pareto set learning method can become trapped on a locally Pareto-optimal component when the true Pareto front is disconnected.

Consider the bi-objective minimization problem illustrated in Figure~\ref{fig:weak_pareto}
\[
\min_{x \in \mathbb{R}} f(x) = \big(f_1(x), f_2(x)\big),
\]
with two objectives \(f_1, f_2 : \mathbb{R} \to \mathbb{R}\). Suppose that the global Pareto front in objective space consists of two disconnected components
\[
\mathrm{PF} = C_1 \cup C_2, \quad C_1 \cap C_2 = \emptyset,
\]
Let \(\mathrm{PS}_1, \mathrm{PS}_2 \subset \mathbb{R}\) be the corresponding sets of decision variables such that
\[
C_i = f(\mathrm{PS}_i), \quad i = 1,2, \qquad
\mathrm{PS} := \mathrm{PS}_1 \cup \mathrm{PS}_2
\]
is the (global) Pareto set. Assume that \(\mathrm{PS}_1\) and \(\mathrm{PS}_2\) areseparated in decision space by a region of dominated solutions: any path from a point in \(\mathrm{PS}_1\) to a point in \(\mathrm{PS}_2\) must passthrough points that worsen at least one objective.

A point \(x^* \in \mathbb{R}\) is called \emph{locally Pareto optimal} if there exists
\(\varepsilon > 0\) such that there is no \(\|x - x^*\| < \epsilon \) with
\begin{equation}
    \begin{gathered}
        f(x) \prec f(x^*) \\
        \text{(i.e., } f_i(x) \le f_i(x^*)\ \forall i, \text{ and } \exists j: f_j(x) < f_j(x^*)\text{)}.
    \end{gathered}
\end{equation}

By construction, every point of \(\mathrm{PS}_1\) is locally Pareto optimal, because any sufficiently small perturbation stays within the basin of attraction of \(\mathrm{PS}_1\) and cannot reach \(\mathrm{PS}_2\) without first worsening an objective.

Now consider a gradient-based multi-objective update of the form
\[
x_{t+1} = x_t - \eta_t \big( \lambda_1^{(t)} \nabla f_1(x_t)
                          + \lambda_2^{(t)} \nabla f_2(x_t) \big),
\]
where \(\lambda^{(t)} = (\lambda_1^{(t)},\lambda_2^{(t)})\) satisfies
\(\lambda_i^{(t)} \ge 0\) and \(\lambda_1^{(t)} + \lambda_2^{(t)} = 1\), and
\(\eta_t > 0\) is a step size. Suppose the initial point \(x_0\) lies in the basin of attraction of \(\mathrm{PS}_1\). Then the iterates \(\{x_t\}\) converge to some \(x^\dagger \in \mathrm{PS}_1\), and at this point, there is no common descent direction:
\[
\nexists d \in \mathbb{R} : \nabla f_1(x^\dagger)^\top d < 0,\quad
                            \nabla f_2(x^\dagger)^\top d < 0.
\]
Thus \(x^\dagger\) is Pareto-stationary and locally Pareto optimal, and any small step towards \(\mathrm{PS}_2\) would increase at least one objective, so the gradient-based method has no incentive to move away from \(x^\dagger\).

In a Pareto set learning model with a finite collection of decision parameters \(\{\lambda_k\}_{k=1}^K\) and a generator \(x_k = \phi_{\theta}(\lambda_k)\), if all initial \(x_k\) lie in the basin of attraction of \(\mathrm{PS}_1\), gradient-based updates will drive all \(x_k\) toward \(\mathrm{PS}_1\). Once each \(x_k\) becomes locally Pareto optimal, the training procedure stops, even though the disconnected component \(\mathrm{PS}_2\) yields strictly better trade-offs. Consequently, the learned Pareto set is \emph{locally} complete around \(\mathrm{PS}_1\) but
\emph{globally} incomplete, since the superior component \(\mathrm{PS}_2\) is never discovered.

\begin{figure*}[!tbhp]
	\centering
	\includegraphics[width=.9\linewidth]{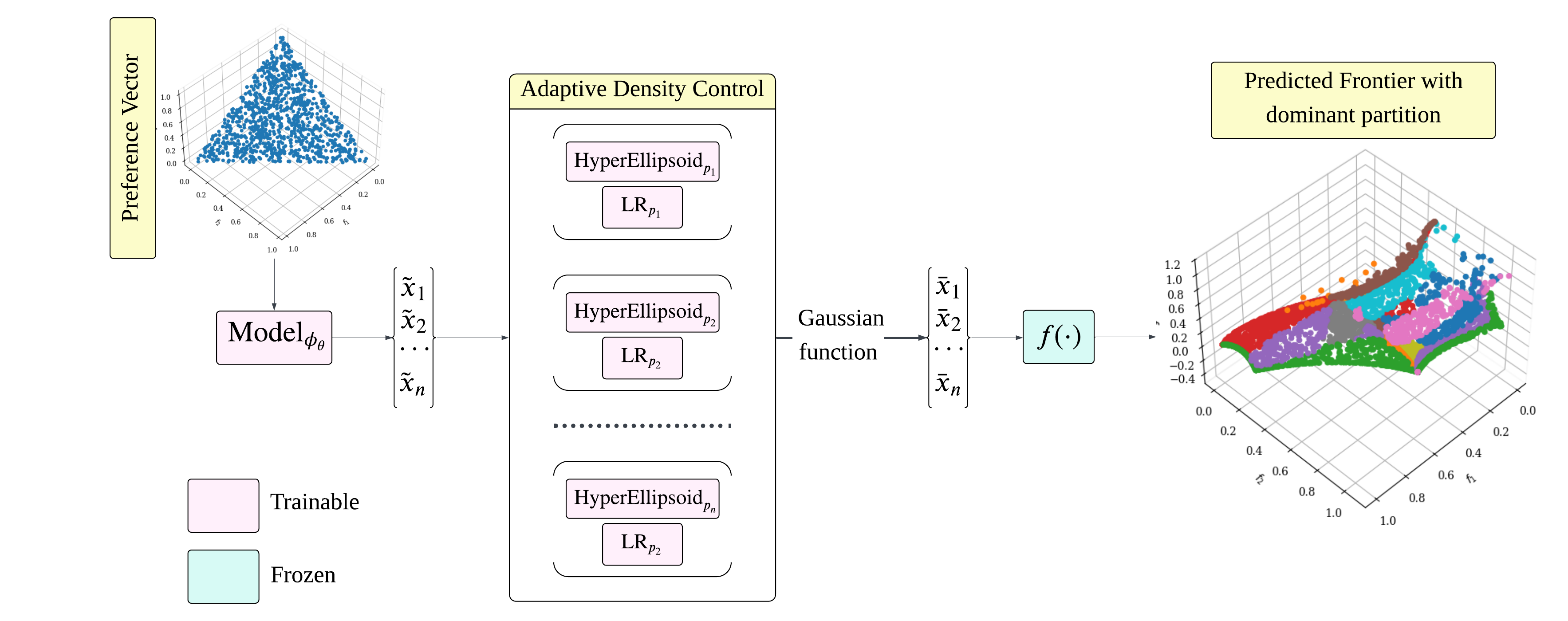}
	\caption{The overall architecture of the framework using the soft partitioning on the decision space}
	\label{fig:framework}
\end{figure*}

\section{Proposed Method: GaussianPSL}

Figure~\ref{fig:framework} illustrates the overall architecture of GaussianPSL, which addresses complex Pareto frontiers via soft partitioning of the decision space, as detailed in the following.
%\subsection{: Soft partitioning on the Pareto space}
\subsection{Soft Partitioning on Decision Space}
Inspired by the Gaussian Splatting framework \cite{kerbl3Dgaussians}, we introduce soft partitioning on the Pareto decision space. To model complex, uneven regions of the Pareto frontier, we partition the latent space using an adaptive density-control mechanism. Each partition is represented by a hyper-ellipsoid and handled by a corresponding local regressor, enabling localized specialization. With these partitions, the PSL model could:
\begin{itemize}
    \item Separating the decision vectors into distinct spatial regions around partition centers
    \item Prevent gradient from collapsing to a single weak Pareto stationary point
    \item Allow each partition to explore different decision space regions independently
\end{itemize}

In 3D space, each ellipsoid could be represented by a Gaussian function as follows:
 \begin{equation} \label{eq:gs}
	\begin{split}
		G_{(\mu,  \Sigma)}(p) &= \sigma(\alpha)e^{-\frac{1}{2}(p-\mu)^T\Sigma^{-1}(p-\mu)} \\
		\Sigma &= RSS^TR^T \\ 
		R &= \begin{bmatrix}
			r_{xx} & r_{xy} & r_{xz} \\
			r_{yx} & r_{yy} & r_{yz} \\
			r_{zx} & r_{yz} & r_{zz}
		\end{bmatrix} \\
		S &= \begin{bmatrix}
			s_x & 0 & 0 \\
			0 & s_y & 0 \\
			0 & 0 & s_z
		\end{bmatrix}
	\end{split}
\end{equation}    
where $\Sigma$ is the covariance matrix of the Gaussian distribution, presented by the 3×3 rotation matrix $R$ analytically expressed with four quaternions and a diagonal scaling matrix $S$ with three parameters for scale. Function $G_{(\mu,  \Sigma)}(p)$ computes the effect of the ellipsoid represented by the Gaussian distribution $N(\mu, \Sigma)$ on the point $p$. This function involves distance from $p$ to the ellipsoid center, and the basic transparency of the ellipsoid $\sigma(\alpha) = \{0..1\}$. The representation of $R$ could be extended to higher dimensions using the generalization of Euler angles \cite{10.1063/1.1666011}. The details of how to generate matrix $\Sigma$ in high dimension is described in Appendix \ref{appendix:gs_nd}.

\subsection{Adaptive Density Control}
\begin{figure}
	\centering
	\includegraphics[width=0.8\linewidth]{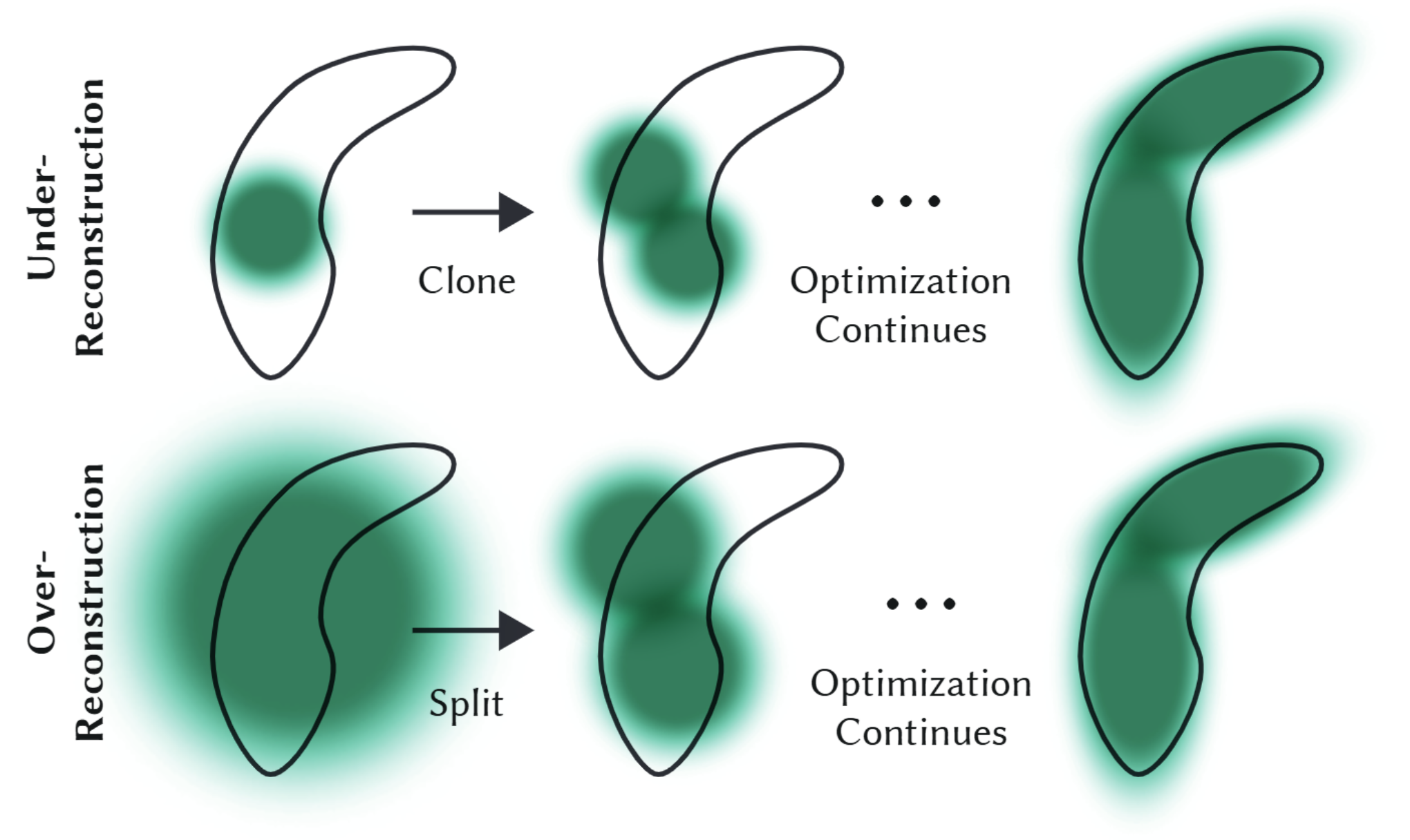}
	\caption{Top row (under-reconstruction): When small-scale geometry (black outline) is insufficiently covered, clone the respective elipsoid. Bottom row (over-reconstruction): If small-scale geometry is represented by one large splat, split it into two. Source: \cite{kerbl3Dgaussians}}
	\label{fig:gssplitting}
\end{figure}

Although constructing a hyperelipsoid ensures differentiability, it is important to note that gradient-based models alone can only adjust existing points; they struggle to discover suitable parameters in regions with insufficient points or with too many points. This is where adaptive densification becomes essential. At first, we randomly inititate the hyperellipsoid. And during training, these hyperellipsoids can move and rotate freely to adapt to the latent space. After several epochs, the model may identify ellipsoids that fail to converge and are continually shifting within a localized area. This behavior typically occurs when the occupied space is not ellipsoidal. To improve the fit, the model may clone or split the ellipsoid into multiple smaller ellipsoids, as illustrated in Fig. \ref{fig:gssplitting}. Transparent ellipsoids (i.e., those for which $\sigma(\alpha) \approx 0$) or too small ($S \approx \mathbf{0}$) are discarded from the model to save resources and keep robustness.

This technique allows the model to cover an unknown space using ellipsoids without prior knowledge of their exact number or position. It can also be regarded as the core principle underlying GS-based methods. Although the loss functions and ellipsoid parameterizations may vary across different GS method variants~\cite{kerbl3Dgaussians, kheradmand20243d, Huang2DGS2024}, the ideal process of Adaptive Density Control remains the same. 

The final output of GaussianPSL is a predicted Pareto frontier, where different colors indicate the dominant partitions, corresponding to the hyper-ellipsoids and local regressors that contribute most strongly in each region. This visualization illustrates how distinct areas of the frontier are governed by different local models, reflecting the adaptive density control mechanism. As a result, the frontier is approximated in a structured, piecewise-smooth manner that captures both global trends and localized variations.

To encourage the model to prioritize local information from the subspaces, we incorporate subspace entropy for each preference vector. The final loss function of the proposed framework is defined as:
\begin{equation}
	\mathcal{L} =  {\mathbb{E}_{\mathbf{\lambda} \sim \mathcal{D}}}\left (s(f(\phi_\mathbf{\theta} (\mathbf{\lambda})) + \gamma H(\phi_\mathbf{\theta}(\lambda)) - \rho C\right) 
\end{equation}
Here, $H(\phi_\mathbf{\theta})$ denotes the entropy computed based on the distance between estimated decision and the centers of the subspaces, $C$ is the minimum spanning of the partition in the objective space, and $\gamma, \rho$ are predefined hyperparameters. A lower value of $H(\phi_\mathbf{\theta}(\lambda))$ indicates larger subspaces with fewer boundaries between them, which mitigates the risk of overparameterization due to an excessive number of subspaces, while a lower value of $C$ indicates less important regions on the frontier.
\section{Experiments}
\subsection{Baseline Methods} 
In this work, we examine several commonly used scalarization strategies: Linear scalarization (LS)~\cite{navon2021learning}, 
COSMOS~\cite{cosmos}, Chebyshev scalarization (TCH)~\cite{lin2022pareto, lin2022expensive}, Modified Chebyshev scalarization (MTCH)~\cite{WANG2020308, 7927726},Hypervolume (HV)~ \cite{hypervolume_definition, zhang2023}. The formulation of these scalarization functions can be found in Appendix~\ref{appendix:preference_vector}.

We compare our method with the vanilla PSL, SVH-PSL \cite{nguyen2025improving}, using a Stein variational gradient descent and EPS~\cite{Ye2024EvolutionaryPS} using an Evolutionary algorithm on the preference vector selection process for completeness. 

Several widely used benchmark problems~\cite{Deb2005, 10.1162/106365600568202} and real-world (RE)~\cite{Tanabe_2020} problems with different shapes of the Pareto fronts and the number of objectives ($m$) are used for testing. In our experiments, RE problems are denoted as RExy, where x is the number of objectives and y is the problem id; e.g., RE21 denotes real-world problems with 2 objectives and id 1, associated with the issues RE2-4-1. The same definition applies to RE33, RE36, and RE37. The Pareto fronts of tested problems can be divided into four types: \textbf{Convex} (RE21 ($m=2, n=4$)), \textbf{Continuous, Non-convex} (RE37 ($m=3, n=4$)), \textbf{Disconnected} (DTLZ7 ($m=3, n=10$), ZDT3 ($m=2, n=10$)), \textbf{Degenerated} (DTLZ5 ($m=3, n=10$), RE36 ($m=3, n=4$)).

\subsection{Evaluation Metric} 
We use Log Hypervolume Difference (LHD) to evaluate the quality of a learned Pareto front, denoted as $\hat{\mathbf{Y}}$, compared to the true approximate Pareto front for the synthetic/real-world problems, represented as $\mathbf{Y^*}$. The calculation of LHD involves taking the logarithm of the difference in hypervolumes between $\hat{\mathbf{Y}}$ and $\mathbf{Y^*}$ as follows:
\begin{equation}
    LHD(\hat{\mathbf{Y}}, \mathbf{Y^*}) = \log \left ( HV(\mathbf{Y^*}) + \epsilon - HV(\hat{\mathbf{Y}}) \right )
\end{equation}
with $\epsilon \in \mathbb{R}_+$ is a small positive value. The smaller the log HV difference, the better the performance.

We also use IGD to evaluate how well a multi-objective optimization algorithm approximates the true Pareto front by measuring both convergence and diversity of the obtained solutions. Lower IGD values indicate better coverage and closeness to the true front. However, since this indicator requires ground truth solutions, it cannot be used for problems where the ground truth is infinite (or very large), such as RE33. The formulation for these indicators could be found in Appendix~\ref{appendix:indicators}.

\subsection{Experiment Settings} 
In all experiments\footnote{The code will be released publicly after the peer review process}, the models (except the SVH-PSL) are trained for $3000$ iterations. All algorithms are evaluated on seven test problems using the same set of $30$ random seeds to ensure robustness. The hyperparameters $\gamma, \rho$ are set to $100$ and $7$, respectively.

\subsection{Experimental Results}
\begin{table*}
\centering
\caption{Log Hypervolume Difference indicator across different problems for models using various scalarization losses (mean with standard deviation)}
\label{table:lhd}
\renewcommand{\arraystretch}{1.15}
\setlength{\tabcolsep}{6pt}
\begin{tabular}{c l c c c c c}
\hline
 &  & \textbf{MTCH} & \textbf{TCH} & \textbf{COSMOS} & \textbf{LS} & \textbf{HV} \\
\hline

\multirow{4}{*}{\textbf{RE21}}
& PSL & -0.691 (0.029) & -0.691 (0.029) & 2.853 (1.019) & -0.642 (0.006) & -0.528 (0.030) \\
& EPS          & -0.258 (0.039) & -0.266 (0.039) & \textbf{0.518 (0.099)} & -0.282 (0.029) & 0.206 (0.132) \\
% & SVH-PSL      & \multicolumn{5}{c}{-0.071 (0.080)} \\
& GaussianPSL         & \textbf{-0.720 (0.030)} & \textbf{-0.711 (0.036)} & 1.484 (0.419) & \textbf{-0.658 (0.005)} & \textbf{-0.651 (0.122)} \\
\hline

\multirow{4}{*}{\textbf{RE33}}
& PSL  & 2.451 (1.103) & 3.197 (0.550) & 2.588 (0.729) & 3.253 (0.569) & 3.234 (0.429) \\
& EPS          & 3.019 (0.284) & 3.238 (0.059) & \textbf{2.114 (0.275)} & 3.284 (0.003) & 3.778 (0.275) \\
% & SVH-PSL      & \multicolumn{5}{c}{1.901 (0.754)} \\
& GaussianPSL         & \textbf{-0.808 (1.356)} & \textbf{0.655 (1.039)} & 2.916 (0.697) & \textbf{1.447 (0.847)} & \textbf{-0.035 (0.883)} \\
\hline

\multirow{4}{*}{\textbf{RE36}}
& PSL  & \textbf{-2.441 (0.548)} & -2.702 (0.475) & 2.220 (1.085) & \textbf{-2.444 (0.329)} & \textbf{-1.446 (0.639)} \\
& EPS          & -1.694 (0.315) & -2.278 (0.207) & 1.044 (0.440) & -2.092 (0.201) & 1.074 (0.643) \\
% & SVH-PSL      & \multicolumn{5}{c}{1.415 (1.082)} \\
& GaussianPSL         & -2.230 (0.261) & \textbf{-2.812 (0.237)} & \textbf{0.158 (0.271)} & -1.072 (0.181) & -0.633 (0.213) \\
\hline

\multirow{4}{*}{\textbf{RE37}}
& PSL  & -4.491 (1.637) & -3.240 (1.813) & -3.218 (0.150) & -1.638 (0.357) & -3.199 (1.469) \\
& EPS          & -4.382 (0.104) & -3.667 (0.079) & -2.373 (0.030) & -2.604 (0.981) & -2.927 (0.670) \\
% & SVH-PSL      & \multicolumn{5}{c}{-3.315 (0.263)} \\
& GaussianPSL         & \textbf{-4.491 (0.828)} & \textbf{-6.875 (0.858)} & \textbf{-3.240 (0.133)} & \textbf{-2.747 (0.390)} & \textbf{-5.597 (0.858)} \\
\hline

\multirow{4}{*}{\textbf{DTLZ5}}
& PSL  & \textbf{-6.671 (0.054)} & \textbf{-7.033 (0.073)} & \textbf{-4.431 (0.143)} & -4.335 (0.534) & \textbf{-6.388 (0.143)} \\
& EPS          & -5.666 (0.059) & -5.544 (0.117) & -3.233 (0.143) & \textbf{-5.629 (0.088)} & -5.200 (0.445) \\
% & SVH-PSL      & \multicolumn{5}{c}{-1.644 (0.512)} \\
& GaussianPSL         & -6.450 (0.061) & -6.854 (0.051) & -3.460 (0.590) & -4.528 (0.472) & -6.379 (0.200) \\
\hline

\multirow{4}{*}{\textbf{DTLZ7}}
& PSL  & -0.802 (0.790) & -0.824 (0.765) & 0.968 (0.068) & -0.045 (0.286) & -1.905 (1.003) \\
& EPS          & -1.741 (0.327) & -1.714 (0.539) & 0.604 (0.009) & -0.824 (0.602) & -0.691 (0.210) \\
% & SVH-PSL      & \multicolumn{5}{c}{-0.234 (0.380)} \\
& GaussianPSL         & \textbf{-2.234 (0.122)} & \textbf{-2.061 (0.048)} & \textbf{0.170 (0.531)} & \textbf{-1.861 (0.152)} & \textbf{-2.555 (0.125)} \\
\hline

\multirow{4}{*}{\textbf{ZDT3}}
& PSL  & -0.853 (0.507) & -0.689 (0.559) & -1.168 (0.566) & -0.585 (0.944) & -5.317 (1.523) \\
& EPS          & -3.841 (1.560) & -4.434 (1.398) & -2.488 (0.604) & \textbf{-5.373 (0.464)} & -5.298 (0.306) \\
% & SVH-PSL      & \multicolumn{5}{c}{-1.154 (0.825)} \\
& GaussianPSL         & \textbf{-4.568 (1.604)} & \textbf{-4.559 (1.337)} & \textbf{-3.142 (0.869)} & -3.845 (1.141) & \textbf{-5.655 (0.541)} \\
\hline

\end{tabular}
\end{table*}

\begin{table*}
\centering
\caption{IGD indicator across different problems for models using various scalarization losses (mean with standard deviation). The RE33 is discarded from the results due to an infinite (too large) objective scale.}
\label{table:igd}
\renewcommand{\arraystretch}{1.15}
\setlength{\tabcolsep}{6pt}
\begin{tabular}{c l c c c c c}
\hline
 &  & \textbf{MTCH} & \textbf{TCH} & \textbf{COSMOS} & \textbf{LS} & \textbf{HV} \\
\hline

\multirow{4}{*}{\textbf{RE21}}
& PSL  & 2.086 (0.056) & 2.083 (0.075) & 291.308 (162.143) & 3.053 (0.038) & \textbf{2.129 (0.132)} \\
& EPS          & 4.590 (1.927) & 4.165 (1.003) & \textbf{46.905 (15.149)} & 8.802 (2.583) & 46.229 (27.217) \\
% & SVH-PSL      & \multicolumn{5}{c}{4.606 (0.191)} \\
& GaussianPSL         & \textbf{2.044 (0.089)} & \textbf{2.068 (0.126)} & 59.482 (37.538) & \textbf{2.885 (0.067)} & 2.220 (0.369) \\
\hline

\multirow{4}{*}{\textbf{RE36}}
& PSL  & 1.960 (0.084) & 1.978 (0.057) & 1.129 (0.841) & 1.976 (0.066) & 1.014 (0.344) \\
& EPS          & 1.824 (0.519) & 1.889 (0.103) & 2.493 (1.973) & 1.905 (0.204) & 21.049 (5.387) \\
% & SVH-PSL      & \multicolumn{5}{c}{2.150 (2.124)} \\
& GaussianPSL         & \textbf{1.658 (0.257)} & \textbf{1.824 (0.058)} & \textbf{0.646 (0.171)} & \textbf{1.474 (0.142)} & \textbf{0.625 (0.177)} \\
\hline

\multirow{4}{*}{\textbf{RE37}}
& PSL  & 0.130 (0.105) & 0.194 (0.098) & 0.060 (0.004) & 0.358 (0.078) & 0.177 (0.069) \\
& EPS          & 0.081 (0.016) & 0.116 (0.008) & 0.095 (0.005) & 0.336 (0.043) & 0.318 (0.029) \\
% & SVH-PSL      & \multicolumn{5}{c}{0.095 (0.012)} \\
& GaussianPSL         & \textbf{0.073 (0.033)} & \textbf{0.063 (0.013)} & \textbf{0.058 (0.003)} & \textbf{0.138 (0.039)} & \textbf{0.082 (0.011)} \\
\hline

\multirow{4}{*}{\textbf{DTLZ5}}
& PSL  & \textbf{0.002 (0.000)} & \textbf{0.002 (0.000)} & \textbf{0.009 (0.001)} & 0.014 (0.009) & \textbf{0.002 (0.000)} \\
& EPS          & 0.002 (0.001) & 0.002 (0.001) & 0.039 (0.005) & 0.248 (0.319) & 0.007 (0.007) \\
% & SVH-PSL      & \multicolumn{5}{c}{0.233 (0.105)} \\
& GaussianPSL         & 0.003 (0.000) & \textbf{0.002 (0.000)} & 0.016 (0.009) & \textbf{0.012 (0.005)} & \textbf{0.002 (0.000)} \\
\hline

\multirow{4}{*}{\textbf{DTLZ7}}
& PSL  & 0.378 (0.222) & 0.393 (0.264) & 1.261 (0.026) & 0.700 (0.218) & 0.163 (0.145) \\
& EPS          & 0.459 (0.313) & 0.397 (0.278) & 0.572 (0.037) & 0.737 (0.156) & 0.660 (0.271) \\
% & SVH-PSL      & \multicolumn{5}{c}{0.518 (0.315)} \\
& GaussianPSL         & \textbf{0.068 (0.004)} & \textbf{0.072 (0.005)} & \textbf{0.335 (0.115)} & \textbf{0.081 (0.016)} & \textbf{0.046 (0.005)} \\
\hline

\multirow{4}{*}{\textbf{ZDT3}}
& PSL  & 0.428 (0.225) & 0.502 (0.233) & 0.262 (0.139) & 0.566 (0.251) & 0.029 (0.039) \\
& EPS          & 0.213 (0.186) & 0.312 (0.171) & 0.210 (0.060) & 0.595 (0.320) & 0.036 (0.020) \\
% & SVH-PSL      & \multicolumn{5}{c}{0.304 (0.229)} \\
& GaussianPSL    & \textbf{0.058 (0.111)} & \textbf{0.053 (0.120)} & \textbf{0.051 (0.050)} & \textbf{0.065 (0.136)} & \textbf{0.012 (0.003)} \\
\hline

\end{tabular}
\end{table*}

\begin{table*}[htbp]
\centering
\caption{Number of essential partition (mean with standard deviation) across test problems}
\label{table:n_part}
\begin{tabular}{lccccc}
\hline
\textbf{Problem} & \textbf{MTCH} & \textbf{TCH} & \textbf{COSMOS} & \textbf{LS} & \textbf{HV} \\
\hline
RE21  & 5.21 (1.78) & 5.29 (1.71) & 4.68 (1.91) & \textbf{4.07 (1.41)} & 5.18 (1.71) \\
RE33  & 6.43 (3.35) & 11.89 (6.59) & \textbf{4.36 (3.18)} & 6.93 (4.00) & 5.61 (3.27) \\
RE36  & 17.14 (6.17) & 15.43 (4.00) & 12.107 (3.35) & \textbf{2.82 (0.76)} & 10.64 (5.85) \\
RE37  & 12.96 (6.38) & 17.39 (4.98) & 9.607 (3.416) & \textbf{4.57 (3.12)} & 20.21 (8.01) \\
DTLZ5 & 15.46 (3.61) & 16.32 (4.98) & 3.893 (1.877) & \textbf{2.39 (0.62)} & 14.57 (5.10) \\
DTLZ7 & 18.11 (5.35) & 7.43 (3.22) & 5.071 (2.999) & \textbf{4.25 (1.64)} & 13.68 (6.09) \\
ZDT3  & 3.04 (1.27) & 3.93 (1.79) & 3.571 (1.347) & \textbf{2.07 (0.46)} & 4.89 (1.97) \\
\hline
\end{tabular}
\end{table*}

\begin{table}[t]
\centering
\caption{SVH-PSL performance summary (mean with standard deviation)}
\label{table:svh-psl}
\renewcommand{\arraystretch}{1.15}
\setlength{\tabcolsep}{10pt}
\begin{tabular}{l c c}
\hline
\textbf{Problem} & \textbf{LHD} & \textbf{IGD} \\
\hline
RE21   & -0.071 (0.080) & 4.606 (0.191) \\
RE33   &  1.901 (0.754) & -- \\
RE36   &  1.415 (1.082) & 2.150 (2.124) \\
RE37   & -3.315 (0.263) & 0.095 (0.012) \\
DTLZ5  & -1.644 (0.512) & 0.233 (0.105) \\
DTLZ7  & -0.234 (0.380) & 0.518 (0.315) \\
ZDT3   & -1.154 (0.825) & 0.304 (0.229) \\
\hline
\end{tabular}
\end{table}

Tables \ref{table:lhd} and \ref{table:igd} report the comparative performance of different PSL-based scalarization strategies using log Hypervolume (HV) and Inverted Generational Distance (IGD), respectively. Lower values indicate better performance for both metrics. Overall, the results show that our method consistently achieves superior or highly competitive performance across a wide range of benchmark problems and Pareto front geometries, often with reduced variance, indicating improved robustness and stability compared to the vanilla and EPS baselines.

For the convex problem RE21, our method demonstrates clear advantages in both convergence and stability. In terms of log HV, our approach achieves the best results for most scalarization strategies, outperforming both vanilla PSL and EPS. The IGD results further confirm this trend, with our method attaining the lowest IGD values under MTCH, TCH, and LS. These results indicate that our method is particularly effective at approximating convex Pareto fronts with high accuracy and consistent performance. Notice that, this problem could have a high IGD indication due to the fact that its objectives reach near 3000. 

The continuous, non-convex problem RE37 highlights the strengths of our method in handling complex Pareto front geometries. Our approach consistently achieves the best log HV across all scalarization strategies, often by a significant margin. This superiority is reinforced by the IGD results, where our method yields the lowest values with small standard deviations across all variants. These outcomes demonstrate the ability of our method to maintain both good diversity and strong convergence in the presence of non-convexity.

For disconnected Pareto fronts, represented by DTLZ7 and ZDT3, our method shows a marked advantage over the baseline approaches. In the log HV results, our method consistently achieves the lowest values across most scalarizations, indicating improved coverage of all disconnected regions. The IGD results corroborate these findings, with our approach substantially reducing the distance to the true Pareto front compared to both vanilla PSL and EPS. This suggests that our method is particularly robust in preserving diversity when the Pareto front is fragmented.

In the case of degenerated problems, namely DTLZ5 and RE36, our method remains competitive and often superior, especially in terms of IGD. While the vanilla PSL occasionally achieves comparable or slightly better HV values on DTLZ5, our method consistently maintains low IGD values with small variance, reflecting an accurate approximation of the reduced-dimensional Pareto front. For RE36, our approach clearly outperforms both baselines in IGD across all scalarization strategies and achieves strong HV performance, particularly under TCH.

Following the standard settings of SVH-PSL \cite{nguyen2025improving}, we also re-run the model on our list of problems. The results are reported in the Table~\ref{table:svh-psl}. Although in the settings of a small number of evaluations, SVH-PSL outperforms the baseline, in our experiments, when the models are freely accessed by the evaluators, with appropriate choices of scalarizations, our GaussianPSL, vanilla PSL, or EPS could easily outperform the SVH-PSL.

\subsection{Ablation study}
We investigate the number of important partitions—those exerting the strongest influence on Pareto front prediction—with results reported in Table \ref{table:n_part}. While MTCH, TCH, and HV scalarizations achieve superior performance compared to COSMOS and LS, they require substantially more partitions. Notably, LS scalarization demands the fewest partitions across most problems while delivering competitive results, despite prior research indicating its limitations for non-convex fronts \cite{convex_optimization, Das1997, Miettinen1999}.

This finding reveals overparameterization in MTCH, TCH, and HV scalarizations (except on the convex RE21 problem), as they utilize more partitions than necessary when benchmarking against the efficient LS baseline. Consequently, our proposed framework requires a more effective mechanism for controlling partition counts to optimize computational resources. Although adaptive density control can increase partitions as needed, the current pruning conditions fail to maintain model compactness.

In summary, across convex, non-convex, disconnected, and degenerate problem categories, our method demonstrates robust optimization performance with consistent HV and IGD improvements, lower run-to-run variability, and strong generalization across scalarization strategies. However, the framework suffers from overparameterization, necessitating refinements to pruning and density control for computational efficiency.

\section{Conclusion}
In this paper, we introduce GaussianPSL, a novel Pareto Set Learning framework that employs soft partitioning of the Pareto decision/objective space to overcome limitations of traditional scalarization and PSL methods on complex frontiers (non-convex, disconnected, degenerate). By enabling simple MLPs to learn localized features within partitions and aggregate them for prediction, GaussianPSL mitigates weak Pareto pollution and local optima trapping while improving exploration, convergence, and initialization robustness. Experiments across diverse problem categories show consistent HV and IGD improvements with lower variability versus baselines, with LS scalarization proving most parameter-efficient despite prior non-convex limitations. However, MTCH, TCH, and HV scalarization exhibit overparameterization requiring excessive partitions, indicating a need for refined adaptive pruning. GaussianPSL establishes a scalable paradigm for robust MOO, with future work targeting a more efficient partitioning mechanism and transfer learning for real-world applications.

\section*{Impact Statement}
This work improves Pareto set learning in the presence of complex and irregular frontiers, enabling more accurate approximation of multi-objective trade-offs where existing methods struggle. By reliably capturing non-convex, discontinuous, and highly structured Pareto fronts, the approach supports robust decision-making in applications such as engineering design, resource allocation, and machine learning model selection.

\bibliography{example_paper}

@inproceedings{navon2021learning,
	title={Learning the Pareto Front with Hypernetworks},
	author={Aviv Navon and Aviv Shamsian and Gal Chechik and Ethan Fetaya},
	booktitle={International Conference on Learning Representations},
	year={2021},
	url={https://openreview.net/forum?id=NjF772F4ZZR}
}

@article{Ye2024EvolutionaryPS,
	title={Evolutionary Preference Sampling for Pareto Set Learning},
	author={Rongguang Ye and Longcan Chen and Jinyuan Zhang and Hisao Ishibuchi},
	journal={Proceedings of the Genetic and Evolutionary Computation Conference},
	year={2024},
	url={https://api.semanticscholar.org/CorpusID:269137143}
}

@inproceedings{zhang2023,
	author = {Zhang, Xiaoyuan and Lin, Xi and Xue, Bo and Chen, Yifan and Zhang, Qingfu},
	title = {Hypervolume maximization: a geometric view of pareto set learning},
	year = {2023},
	publisher = {Curran Associates Inc.},
	address = {Red Hook, NY, USA},
	booktitle = {Proceedings of the 37th International Conference on Neural Information Processing Systems},
	articleno = {1690},
	numpages = {28},
	location = {New Orleans, LA, USA},
	series = {NIPS '23}
}

@inproceedings{lin2022pareto,
	title={Pareto Set Learning for Neural Multi-Objective Combinatorial Optimization},
	author={Xi Lin, Zhiyuan Yang, Qingfu Zhang},
	booktitle={International Conference on Learning Representations},
	year={2022},
	url={https://openreview.net/forum?id=QuObT9BTWo}
}

@inproceedings{lin2022expensive,
	author = {Lin, Xi and Yang, Zhiyuan and Zhang, Xiaoyuan and Zhang, Qingfu},
	title = {Pareto set learning for expensive multi-objective optimization},
	year = {2022},
	isbn = {9781713871088},
	publisher = {Curran Associates Inc.},
	address = {Red Hook, NY, USA},
	booktitle = {Proceedings of the 36th International Conference on Neural Information Processing Systems},
	articleno = {1398},
	numpages = {17},
	location = {New Orleans, LA, USA},
	series = {NIPS '22}
}

@INPROCEEDINGS{cosmos,
  author={Ruchte, Michael and Grabocka, Josif},
  booktitle={2021 IEEE International Conference on Data Mining (ICDM)}, 
  title={Scalable Pareto Front Approximation for Deep Multi-Objective Learning}, 
  year={2021},
  volume={},
  number={},
  pages={1306-1311},
  doi={10.1109/ICDM51629.2021.00162}}

@book{convex_optimization, 
	place={Cambridge},
	title={Convex Optimization}, 
	publisher={Cambridge University Press}, 
	author={Boyd, Stephen and Vandenberghe, Lieven}, 
	year={2004}
}

@InProceedings{pmlr-v119-mahapatra20a,
	title = 	 {Multi-Task Learning with User Preferences: Gradient Descent with Controlled Ascent in Pareto Optimization},
	author =       {Mahapatra, Debabrata and Rajan, Vaibhav},
	booktitle = 	 {Proceedings of the 37th International Conference on Machine Learning},
	pages = 	 {6597--6607},
	year = 	 {2020},
	editor = 	 {III, Hal Daumé and Singh, Aarti},
	volume = 	 {119},
	series = 	 {Proceedings of Machine Learning Research},
	month = 	 {13--18 Jul},
	publisher =    {PMLR},
	url = 	 {https://proceedings.mlr.press/v119/mahapatra20a.html}
}

@ARTICLE{hypervolume_definition,
	author={Zitzler, E. and Thiele, L.},
	journal={IEEE Transactions on Evolutionary Computation}, 
	title={Multiobjective evolutionary algorithms: a comparative case study and the strength Pareto approach}, 
	year={1999},
	volume={3},
	number={4},
	pages={257-271},
	doi={10.1109/4235.797969}}

@InProceedings{IGD,
	author="Coello Coello, Carlos A. and Reyes Sierra, Margarita",
	editor="Monroy, Ra{\'u}l and Arroyo-Figueroa, Gustavo and Sucar, Luis Enrique and Sossa, Humberto",
	title="A Study of the Parallelization of a Coevolutionary Multi-objective Evolutionary Algorithm",
	booktitle="MICAI 2004: Advances in Artificial Intelligence",
	year="2004",
	publisher="Springer Berlin Heidelberg",
	address="Berlin, Heidelberg",
	pages="688--697",
	isbn="978-3-540-24694-7"
}

@Article{kerbl3Dgaussians,
	author       = {Kerbl, Bernhard and Kopanas, Georgios and Leimk{\"u}hler, Thomas and Drettakis, George},
	title        = {3D Gaussian Splatting for Real-Time Radiance Field Rendering},
	journal      = {ACM Transactions on Graphics},
	number       = {4},
	volume       = {42},
	month        = {July},
	year         = {2023},
	url          = {https://repo-sam.inria.fr/fungraph/3d-gaussian-splatting/}
}

@inproceedings{kheradmand20243d,
	title = {3D Gaussian Splatting as Markov Chain Monte Carlo},
	author = {Kheradmand, Shakiba and Rebain, Daniel and Sharma, Gopal and Sun, Weiwei and Tseng, Yang-Che and Isack, Hossam and Kar, Abhishek and Tagliasacchi, Andrea and Yi, Kwang Moo},
	booktitle = {Advances in Neural Information Processing Systems (NeurIPS)},
	year = {2024},
	note = {Spotlight Presentation},
}

@article{Tanabe_2020,
	title={An easy-to-use real-world multi-objective optimization problem suite},
	volume={89},
	ISSN={1568-4946},
	url={http://dx.doi.org/10.1016/j.asoc.2020.106078},
	DOI={10.1016/j.asoc.2020.106078},
	journal={Applied Soft Computing},
	publisher={Elsevier BV},
	author={Tanabe, Ryoji and Ishibuchi, Hisao},
	year={2020},
	month=apr, pages={106078} }

@book{Miettinen1999,
  address = {Boston, USA},
  author = {Miettinen, K.},
  publisher = {Kluwer},
  title = {Nonlinear multiobjective optimization},
  username = {dalbem},
  year = 1999
}

@article{Rai01012003,
    author = {Rahul Rai and Venkat Allada},
    title = {Modular product family design: Agent-based Pareto-optimization and quality loss function-based post-optimal analysis},
    journal = {International Journal of Production Research},
    volume = {41},
    number = {17},
    pages = {4075--4098},
    year = {2003},
    publisher = {Taylor \& Francis},
    doi = {10.1080/0020754031000149248},
    URL = { https://doi.org/10.1080/0020754031000149248},
    eprint = {https://doi.org/10.1080/0020754031000149248}
}

@Article{Sinha2018,
    author={Sinha, Kaushik
    and Suh, Eun Suk},
    title={Pareto-optimization of complex system architecture for structural complexity and modularity},
    journal={Research in Engineering Design},
    year={2018},
    month={Jan},
    day={01},
    volume={29},
    number={1},
    pages={123-141},
    issn={1435-6066},
    doi={10.1007/s00163-017-0260-9},
    url={https://doi.org/10.1007/s00163-017-0260-9}
}

@Inbook{Deb2005,
author="Deb, Kalyanmoy
and Thiele, Lothar
and Laumanns, Marco
and Zitzler, Eckart",
editor="Abraham, Ajith
and Jain, Lakhmi
and Goldberg, Robert",
title="Scalable Test Problems for Evolutionary Multiobjective Optimization",
bookTitle="Evolutionary Multiobjective Optimization: Theoretical Advances and Applications",
year="2005",
publisher="Springer London",
address="London",
pages="105--145",
isbn="978-1-84628-137-2",
doi="10.1007/1-84628-137-7_6",
url="https://doi.org/10.1007/1-84628-137-7_6"
}

@inproceedings{Huang2DGS2024,
    title={2D Gaussian Splatting for Geometrically Accurate Radiance Fields},
    author={Huang, Binbin and Yu, Zehao and Chen, Anpei and Geiger, Andreas and Gao, Shenghua},
    publisher = {Association for Computing Machinery},
    booktitle = {SIGGRAPH 2024 Conference Papers},
    year      = {2024},
    doi       = {10.1145/3641519.3657428}
}

@book{Pareto:1971,
  address = {New York},
  author = {Pareto, Vilfredo},
  booktitle = {Manual of Political Economy (Manuale Di Economia Politica)},
  note = {Translated by Ann S. Schwier and Alfred N. Page},
  pages = {xii, 504 p.},
  publisher = {Kelley},
  title = {Manual of political economy (manuale di economia politica)},
  translator = {Ann S. Schwier and Alfred N. Page},
  year = {1906}
}

@article{LIU2025119541,
title = {Pareto-guided active learning for accelerating surrogate-assisted multi-objective optimization of arch dam shape},
journal = {Engineering Structures},
volume = {326},
pages = {119541},
year = {2025},
issn = {0141-0296},
doi = {https://doi.org/10.1016/j.engstruct.2024.119541},
url = {https://www.sciencedirect.com/science/article/pii/S0141029624021035},
author = {Rui Liu and Gang Ma and Fanhui Kong and Zhitao Ai and Kun Xiong and Wei Zhou and Xiaomao Wang and Xiaolin Chang}
}

@misc{huang2025optimizationweaklyparetoset,
      title={Optimization over the weakly Pareto set and multi-task learning}, 
      author={Lei Huang and Jiawang Nie and Jiajia Wang},
      year={2025},
      eprint={2504.00257},
      archivePrefix={arXiv},
      primaryClass={math.OC},
      url={https://arxiv.org/abs/2504.00257}, 
}

@misc{chen2025,
      title={Gradient-Based Multi-Objective Deep Learning: Algorithms, Theories, Applications, and Beyond}, 
      author={Weiyu Chen and Xiaoyuan Zhang and Baijiong Lin and Xi Lin and Han Zhao and Qingfu Zhang and James T. Kwok},
      year={2025},
      eprint={2501.10945},
      archivePrefix={arXiv},
      primaryClass={cs.LG},
      url={https://arxiv.org/abs/2501.10945}, 
}

@article{SU2024101605,
title = {Fast Pareto set approximation for multi-objective flexible job shop scheduling via parallel preference-conditioned graph reinforcement learning},
journal = {Swarm and Evolutionary Computation},
volume = {88},
pages = {101605},
year = {2024},
issn = {2210-6502},
doi = {https://doi.org/10.1016/j.swevo.2024.101605},
url = {https://www.sciencedirect.com/science/article/pii/S2210650224001433},
author = {Chupeng Su and Cong Zhang and Chuang Wang and Weihong Cen and Gang Chen and Longhan Xie}
}

@Article{Das1997,
author={Das, I.
and Dennis, J. E.},
title={A closer look at drawbacks of minimizing weighted sums of objectives for Pareto set generation in multicriteria optimization problems},
journal={Structural optimization},
year={1997},
month={Aug},
day={01},
volume={14},
number={1},
pages={63-69},
issn={1615-1488},
doi={10.1007/BF01197559},
url={https://doi.org/10.1007/BF01197559}
}

@InProceedings{pmlr-v119-ma20a,
  title = 	 {Efficient Continuous Pareto Exploration in Multi-Task Learning},
  author =       {Ma, Pingchuan and Du, Tao and Matusik, Wojciech},
  booktitle = 	 {Proceedings of the 37th International Conference on Machine Learning},
  pages = 	 {6522--6531},
  year = 	 {2020},
  editor = 	 {III, Hal Daumé and Singh, Aarti},
  volume = 	 {119},
  series = 	 {Proceedings of Machine Learning Research},
  month = 	 {13--18 Jul},
  publisher =    {PMLR},
  url = 	 {https://proceedings.mlr.press/v119/ma20a.html}
}

@article{LI2023101253,
title = {Multimodal multi-objective optimization: Comparative study of the state-of-the-art},
journal = {Swarm and Evolutionary Computation},
volume = {77},
pages = {101253},
year = {2023},
issn = {2210-6502},
doi = {https://doi.org/10.1016/j.swevo.2023.101253},
url = {https://www.sciencedirect.com/science/article/pii/S2210650223000275},
author = {Wenhua Li and Tao Zhang and Rui Wang and Shengjun Huang and Jing Liang}
}

@article{WANG2020308,
	title = {A survey of decomposition approaches in multiobjective evolutionary algorithms},
	journal = {Neurocomputing},
	volume = {408},
	pages = {308-330},
	year = {2020},
	issn = {0925-2312},
	doi = {https://doi.org/10.1016/j.neucom.2020.01.114},
	url = {https://www.sciencedirect.com/science/article/pii/S0925231220304926},
	author = {Jia Wang and Yuchao Su and Qiuzhen Lin and Lijia Ma and Dunwei Gong and Jianqiang Li and Zhong Ming}
}

@ARTICLE{7927726,
	author={Ma, Xiaoliang and Zhang, Qingfu and Tian, Guangdong and Yang, Junshan and Zhu, Zexuan},
	journal={IEEE Transactions on Evolutionary Computation}, 
	title={On Tchebycheff Decomposition Approaches for Multiobjective Evolutionary Optimization}, 
	year={2018},
	volume={22},
	number={2},
	pages={226-244},
	doi={10.1109/TEVC.2017.2704118}}

@misc{zheng2025weakparetoboundaryachilles,
      title={Weak Pareto Boundary: The Achilles' Heel of Evolutionary Multi-Objective Optimization}, 
      author={Ruihao Zheng and Jingda Deng and Zhenkun Wang},
      year={2025},
      eprint={2505.13854},
      archivePrefix={arXiv},
      primaryClass={cs.NE},
      url={https://arxiv.org/abs/2505.13854}, 
}

@article{article_2,
	author = {Low, Andre and Vissol-Gaudin, Eleonore and Lim, Yee-Fun and Hippalgaonkar, Kedar},
	year = {2023},
	month = {05},
	pages = {11},
	title = {Mapping pareto fronts for efficient multi-objective materials discovery},
	volume = {3},
	journal = {Journal of Materials Informatics},
	doi = {10.20517/jmi.2023.02}
}

@article{10.1063/1.1666011,
    author = {Hoffman, David K. and Raffenetti, Richard C. and Ruedenberg, Klaus},
    title = {Generalization of Euler Angles to N‐Dimensional Orthogonal Matrices},
    journal = {Journal of Mathematical Physics},
    volume = {13},
    number = {4},
    pages = {528-533},
    year = {1972},
    month = {04},
    issn = {0022-2488},
    doi = {10.1063/1.1666011},
    url = {https://doi.org/10.1063/1.1666011},
    eprint = {https://pubs.aip.org/aip/jmp/article-pdf/13/4/528/19147026/528_1_online.pdf},
}

@inproceedings{nguyen2025improving,
  title={Improving pareto set learning for expensive multi-objective optimization via stein variational hypernetworks},
  author={Nguyen, Minh-Duc and Dinh, Phuong Mai and Nguyen, Quang-Huy and Hoang, Long P and Le, Dung D},
  booktitle={Proceedings of the AAAI Conference on Artificial Intelligence},
  volume={39},
  number={18},
  pages={19677--19685},
  year={2025}
}

@inproceedings{Ye2021ParetoNG,
  title={Pareto Navigation Gradient Descent: a First-Order Algorithm for Optimization in Pareto Set},
  author={Mao Ye and Qiang Liu},
  booktitle={Conference on Uncertainty in Artificial Intelligence},
  year={2021},
  url={https://api.semanticscholar.org/CorpusID:239016760}
}

@article{10.1162/106365600568202,
    author = {Zitzler, Eckart and Deb, Kalyanmoy and Thiele, Lothar},
    title = {Comparison of Multiobjective Evolutionary Algorithms: Empirical Results},
    journal = {Evolutionary Computation},
    volume = {8},
    number = {2},
    pages = {173-195},
    year = {2000},
    month = {06},
    issn = {1063-6560},
    doi = {10.1162/106365600568202},
    url = {https://doi.org/10.1162/106365600568202},
    eprint = {https://direct.mit.edu/evco/article-pdf/8/2/173/1493199/106365600568202.pdf},
}
\bibliographystyle{icml2026}

%%%%%%%%%%%%%%%%%%%%%%%%%%%%%%%%%%%%%%%%%%%%%%%%%%%%%%%%%%%%%%%%%%%%%%%%%%%%%%%
%%%%%%%%%%%%%%%%%%%%%%%%%%%%%%%%%%%%%%%%%%%%%%%%%%%%%%%%%%%%%%%%%%%%%%%%%%%%%%%
% APPENDIX
%%%%%%%%%%%%%%%%%%%%%%%%%%%%%%%%%%%%%%%%%%%%%%%%%%%%%%%%%%%%%%%%%%%%%%%%%%%%%%%
%%%%%%%%%%%%%%%%%%%%%%%%%%%%%%%%%%%%%%%%%%%%%%%%%%%%%%%%%%%%%%%%%%%%%%%%%%%%%%%
\newpage
\appendix
\onecolumn
\section{Preference vectors}
\label{appendix:preference_vector}
\subsection{Preference vectors}
Preference vectors are conditioning variables that specify how different objectives should be weighted or traded off, guiding the model to generate solutions corresponding to different regions of the Pareto front.

Concretely, a preference vector is typically a non-negative vector (often normalized to lie on a simplex) whose components represent the relative importance assigned to each objective. During training, the PSL model takes a preference vector as part of its input and learns a mapping from this vector to a decision variable that optimizes the corresponding scalarized objective. By varying the preference vector, the model can produce a diverse set of Pareto-optimal (or near-optimal) solutions, collectively approximating the Pareto set.

\begin{figure}[H]
    \centering
    \includegraphics[width=\linewidth]{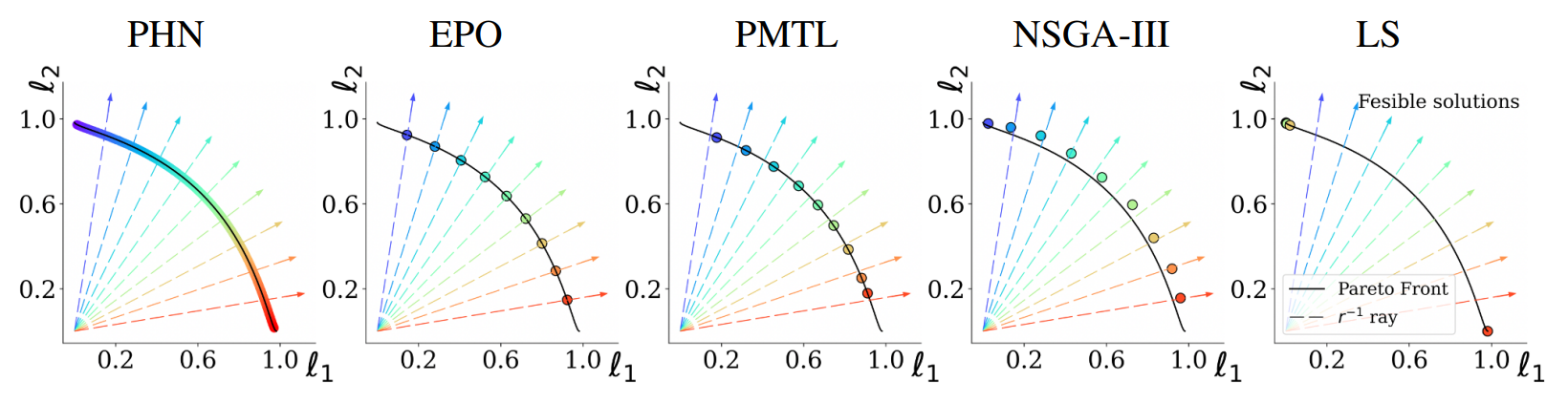}
    \caption{Illustrative example demonstrating the relation between Pareto front, preference rays, and solutions. Source: \cite{navon2021learning}}
    \label{fig:scalarization}
\end{figure}

The selection of the scalarization function will reflect various ways in which the preference vectors intersect the estimated Pareto frontiers. Fig.~\ref{fig:scalarization} from \cite{navon2021learning} illustrates different mappings associated with the preference vectors and choice of scalarization techniques. Pareto front (black solid line) for a 2D loss space and several rays (colored dashed lines), which represent various possible preferences

\subsection{Scalarization loss function}
\label{appendix:scalarization}
Remind the function of PSL model that: 
\begin{equation} \label{eq:psl_opt}
	\begin{split}
		\theta^* &= \argmin_\mathbf{\theta} \mathbb{E}_{\mathbf{\lambda} \sim \mathcal{D}} [s(f(\phi_\mathbf{\theta} (\mathbf{\lambda}))] \\ &= \argmin_\mathbf{\theta} \mathbb{E}_{\mathbf{\lambda} \sim \mathcal{D}} [s(f(\mathbf{x}_\mathbf{\lambda}))]
	\end{split}	
\end{equation}
map $\mathbf{\lambda}$ to a decision vector $\mathbf{x}_\lambda$ such that the resulting objective vector $f(\mathbf{x}_\lambda)$ lies on or near the Pareto frontier.

Formulation for some common choices could be listed as:
\begin{enumerate}
	\item Linear Scalarization (LS) \cite{navon2021learning}. In this method, the optimization function is 
	\begin{equation}  \label{eq:psl_ls}
		\theta^* = \argmin_\mathbf{\theta} \mathbb{E}_{\mathbf{\lambda} \sim Unif(\Delta^{m-1})} \left[ \sum_{i=1}^m \lambda_i f_i(\mathbf{x}_\lambda) \right]
	\end{equation}
	where $Unif(\Delta^{m-1})$ is $(m-1)$-simplex representing by the functions $\sum_i^m \lambda_i = 1$. However, the multiple convex optimization theory research \cite{convex_optimization, Miettinen1999, Das1997} has shown that optimizing Equation (\ref{eq:psl_ls}) can only find the convex part of the Pareto front.
	
	\item COSMOS \cite{cosmos}. Based on the Linear Scalarization loss function, a penalty term was proposed to force the solution
	to obey to the preference vector $\lambda$ by minimizing the angle between $\lambda$ and the vector of losses. Such a desired effect can be modeled by maximizing the cosine similarity between the preference vector and the vector of losses:
	\begin{equation}  \label{eq:psl_cosmos}
		\begin{split}
			\mathcal{L}(\mathbf{\lambda}, \mathbf{x}_\lambda, f(\mathbf{x})) &= \sum_{i=1}^m \lambda_i f_i(\mathbf{x}_\lambda) \\
			\theta^* = \argmin_\mathbf{\theta} \mathbb{E}_{\mathbf{\lambda} \sim Unif(\Delta^{m-1})} & \mathcal{L}(\mathbf{\lambda}, \mathbf{x}_\lambda, f(\mathbf{x}))  \\ &- \tau \frac{\lambda \mathcal{L}(\mathbf{\lambda}, \mathbf{x}_\lambda, f(\mathbf{x}))}{\lVert\lambda\rVert  \lVert\mathcal{L}(\mathbf{\lambda}, \mathbf{x}_\lambda, f(\mathbf{x}))\rVert}
		\end{split}
	\end{equation}
	
	\item Tchebyshev Scalarization (TCH) \cite{lin2022expensive, lin2022pareto}, optimizing: 
		\begin{equation}  \label{eq:psl_tche2}
			\theta^* = \argmin_\mathbf{\theta} \mathbb{E}_{\mathbf{\lambda} \sim Unif(\Delta^{m-1})} \left[ \max_{i \in \{1,\ldots,m\}} \lambda_i (f_i(\mathbf{x}_\lambda) - (\mathbf{z_i^*} -\epsilon)) \right]
		\end{equation}
		where the $\mathbf{z_i^*} = min_{x \in X0 f_i(\mathbf{x})}$ is an ideal value (lower-bound for minimization problem) for each objectives and $\epsilon \in \mathbb{R}_{+}$ is a small value, called an (unachievable) utopia value for the $i$-th objective. This scalarization has an interesting property: "All Pareto solutions can be found by solving the Chebycheff scalarized subproblem with a specific (but unknown) trade-off preference $\lambda$"~\cite{lin2022pareto}. This property is not limited by the geometrical shape of the frontiers, making this scalarization a suitable candidate for the PSL model.  
		
	\item Modified Tchebyshev Scalarization (MTCH)~\cite{WANG2020308, 7927726}, optimizing: 
	\begin{equation}  \label{eq:psl_tche1}
		\theta^* = \argmin_\mathbf{\theta} \mathbb{E}_{\mathbf{\lambda} \sim Unif(\Delta^{m-1})} \left[ \max_{i \in \{1,\ldots,m\}} \frac{1}{\lambda_i} (f_i(\mathbf{x}_\lambda) - (\mathbf{z_i^*} -\epsilon)) \right]
	\end{equation}
	where the $\mathbf{z_i^*} = min_{x \in X0 f_i(\mathbf{x})}$ is and ideal value (lower-bound for minimization problem) for each objectives and $\epsilon \in \mathbb{R}_{+}$ is a small value, called an (unachievable) utopia value for the $i$-th objective. This modified version of vanilla TCH was proposed to obtain the solution at the intersection of the preference vector and the Pareto front.
	
 	\item Hypervolume (HV)~\cite{hypervolume_definition, zhang2023}. This scalarization method is inspired by the evaluator hypervolume in multi-objective space. The hypervolume indicator \cite{hypervolume_definition} is used to measure the volume in the loss space of points dominated by a solution in the evaluated set. As this volume is unbounded, it is restricted to the volume in a rectangle defined by the solutions and a selected reference point. All solutions need to be bounded by the reference point, as illustrated in the Figure \ref{fig:hypervolumeexample}. 
	\begin{figure}
		\centering
		\includegraphics[width=0.4\linewidth]{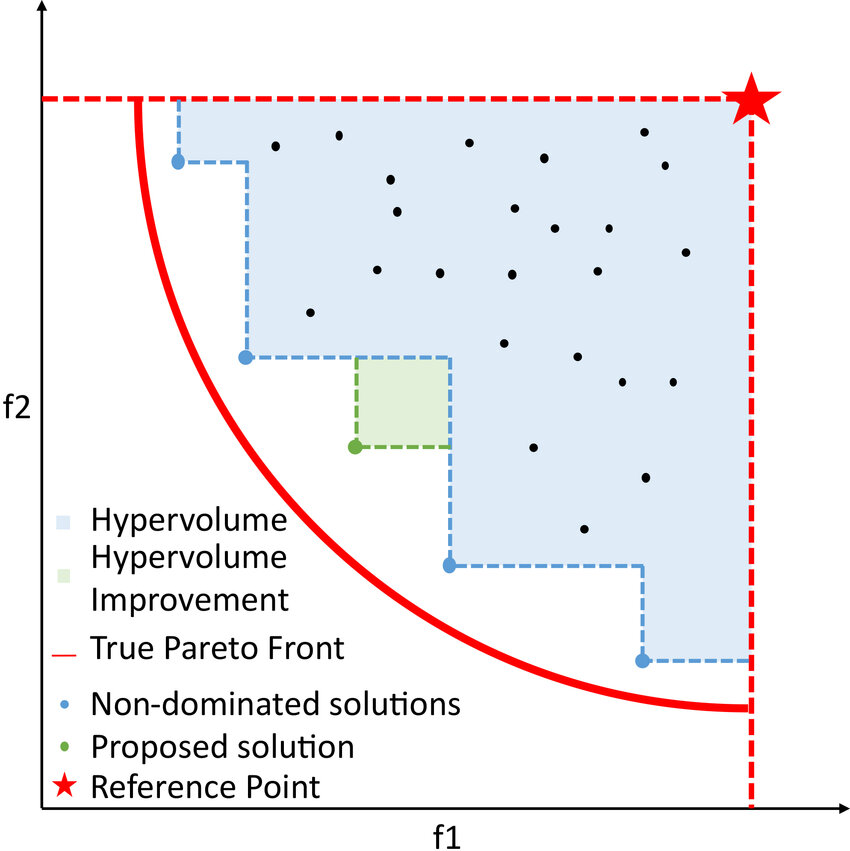}
		\caption{Example of hyper volume in 2D. Source: \cite{article_2}}
		\label{fig:hypervolumeexample}
	\end{figure}
	The actual hyper volume of the Pareto frontier is the yellow area and approximated with the dashed blue area. In a problem where the reference point is outside of the convex hull in the Fig \ref{fig:hypervolumeexample}, the two objectives frontier creates a convex hull; therefore, the optimization function will be:
	\begin{equation} \label{eq:psl_hv}
		\theta^* = \argmin_\mathbf{\theta} HV(\mathbb{E}_{\mathbf{\lambda} \sim Unif(\Delta^{k-1})}f(\mathbf{x}_\lambda))
	\end{equation}
	Due to the nature of the HV, the evaluation needs to compute the Pareto order for each $\hat{\mathbf{y}}$ to identify the points lying on the predicted frontier. This process is non-derivative, which makes it unsuitable for the loss function in a neural network. However,~\citet{zhang2023} had shown that HV could be approximated using a min-max function. 
\end{enumerate}

\section{Evaluation Indicator}
\label{appendix:indicators}
\subsection{Hypervolume Reference Point}
The reference point is set 10\% beyond the nadir of the true Pareto front. The exact value for each problem could be found in the project code. 
\subsection{Inverted Generational Distance}
The Inverted Generational Distance (IGD) \cite{IGD} inverts the generational distance and measures the distance from any point in ${\mathbf{Y}}$ to the closest point in $\hat{\mathbf{y}}^i$. 
\begin{equation} \label{eq:evl_igd}
    IGD(\hat{\mathbf{Y}}) = \frac{1}{N} \left( \sum_{i=1}^{N} \hat{d_i}^{2} \right)^\frac{1}{2}
\end{equation}
where $\hat{d}_i$ represents the Euclidean distance from ${\mathbf{y}^i}$ to its nearest reference point in $\hat{\mathbf{Y}}$.

\section{Extension for Gaussian Splatting in high dimensions}
\label{appendix:gs_nd}
Without the loss for generality, assume matrix $R(\theta)$ performs rotation a counterclockwise angle $\theta$ from basis $E = [e_1, e_2, \dots, e_m]$ to the basis $E' = [e'_1, e'_2, \dots, e'_m]$. Since $E$ and $E'$ are basis matrices, they both satisfy the condition of the Kronecker delta: 
\begin{equation}
    \begin{split}
        \delta_{ij} = e_i \times e_j = \begin{cases}
                                        1 &  i=j \\
                                        0 & i \neq j \\
                                      \end{cases},  \qquad
        \delta'_{ij} = e'_i \times e'_j = \begin{cases}
                                        1 &  i=j \\
                                        0 & i \neq j \\
                                      \end{cases} \\                              
    \end{split}
\end{equation}
Since $R(\theta)$ is a matrix rotation, therefore:
\begin{equation} \label{eq:r_ij}
    \begin{split}
        &E' = R(\theta) \times E \\
        &\Rightarrow e'_i = \sum_{k \in \{1..m\}} R_{ik} \times e_k \\
        &\Rightarrow e'_i \times e_j = \sum_{k \in \{1..m\}} R_{ik} \times e_k \times e_j \\ 
        &\Rightarrow e'_i \times e_j = \sum_{k \in \{1..m\}} R_{ik} \times {e_k \times e_j} \\
        &\Rightarrow e'_i \times e_j = \sum_{k \in \{1..m\}} R_{ik} \times \delta_{kj} \\
        &\Rightarrow e'_i \times e_j =  R_{ij}
    \end{split}
\end{equation}
If $E = I_m$ with $I_m = \mathbbold{1}$ is an identity matrix size $m$, a rotation with angle $\theta$ at the plane contains 2 axises $i$-th and $j$-th ($i \neq j$) could be written in fomular as:
\begin{equation}
    e'_i = \cos(\theta) e_i + \sin(\theta)e_j \qquad
    e'_j = -\sin(\theta) e_i + \cos(\theta)e_j \qquad
    e'_k = e_k \quad k \neq {i, j}
\end{equation}

Based on Equation \ref{eq:r_ij}, the rotation matrix $R(\theta)$ could be computed as:
% \begin{align}
% R_{xy} (x, y \neq i, j) &= e'_x \times e_y \nonumber\\
%                       &= e_x \times e_y = \delta_{xy} \\
% R_{ki} &= e'_k \times e_i \nonumber\\
%        &= e_k \times e_i = \delta_{ki} = 0 \\
% R_{kj} &= e'_k \times e_j \nonumber\\
%        &= e_k \times e_j = \delta_{kj} = 0 \\
% R_{ik} &= e'_i \times e_k \nonumber\\
%        &= \cos(\theta) e_i e_k + \sin(\theta) e_j e_k = 0 \\
% R_{jk} &= e'_j \times e_k \nonumber\\
%        &= -\sin(\theta) e_i e_k + \cos(\theta) e_j e_k = 0 \\
% R_{ij} &= e'_i \times e_j \nonumber\\
%        &= \cos(\theta) \delta_{ij} + \sin(\theta) \delta_{jj} \nonumber\\
%        &= \sin(\theta) \\
% R_{ii} &= e'_i \times e_i \nonumber\\
%        &= \cos(\theta) \delta_{ii} + \sin(\theta) \delta_{ji} \nonumber\\
%        &= \cos(\theta) \\
% R_{jj} &= e'_j \times e_j \nonumber\\
%        &= -\sin(\theta) \delta_{ij} + \cos(\theta) \delta_{jj} \nonumber\\
%        &= \cos(\theta) \\
% R_{ji} &= e'_j \times e_i \nonumber\\
%        &= -\sin(\theta) \delta_{ii} + \cos(\theta) \delta_{ji} \nonumber\\
%        &= -\sin(\theta)
% \end{align}

\noindent
\begin{minipage}{0.48\textwidth}
\begin{align}
\mathop{R_{xy}}\limits_{\substack{x,y\neq i,j}} &= e'_x \times e_y \nonumber\\
                      &= e_x \times e_y = \delta_{xy} \\
R_{ki} &= e'_k \times e_i \nonumber\\
       &= e_k \times e_i = \delta_{ki} = 0 \\
R_{kj} &= e'_k \times e_j \nonumber\\
       &= e_k \times e_j = \delta_{kj} = 0 \\
R_{ik} &= e'_i \times e_k \nonumber\\
       &= \cos(\theta) e_i e_k + \sin(\theta) e_j e_k = 0 \\
R_{jk} &= e'_j \times e_k \nonumber\\
       &= -\sin(\theta) e_i e_k + \cos(\theta) e_j e_k = 0
\end{align}
\end{minipage}
\hfill
\begin{minipage}{0.48\textwidth}
\begin{align}
R_{ij} &= e'_i \times e_j \nonumber\\
       &= \cos(\theta) \delta_{ij} + \sin(\theta) \delta_{jj} \nonumber\\
       &= \sin(\theta) \\
R_{ii} &= e'_i \times e_i \nonumber\\
       &= \cos(\theta) \delta_{ii} + \sin(\theta) \delta_{ji} \nonumber\\
       &= \cos(\theta) \\
R_{jj} &= e'_j \times e_j \nonumber\\
       &= -\sin(\theta) \delta_{ij} + \cos(\theta) \delta_{jj} \nonumber\\
       &= \cos(\theta) \\
R_{ji} &= e'_j \times e_i \nonumber\\
       &= -\sin(\theta) \delta_{ii} + \cos(\theta) \delta_{ji} \nonumber\\
       &= -\sin(\theta)
\end{align}
\end{minipage}

Assuming $R^m \subset \mathbb{R}^{ m \times m} $ is the rotation matrix for hyper-ellipsoids in $m$-D, associated with $m$ objectives. Without the loss of the orthonormality of the rotation matrix $R^m$, we defactorize $R^m$ into multiple rotations in 2D at a plane containing 2 arbitrary axes of the original basis:
\begin{equation} \label{eq:r_m}
    R^m = \prod_{{i, j} \in \binom{m}{2}} R^{ij}(\theta_{ij})
\end{equation}
with $R^{ij}(\theta_{ij})$ is the rotation matrix for the plane contains 2 axises $i$-th and $j$-th to $\theta_{ij}$ angles. Hoffman's research \cite{10.1063/1.1666011} indicated that in four dimensions and higher, the idea of "rotation about an axis" becomes ambiguous and is instead interpreted as "rotation within a plane." The number of Euler angles required to represent the rotation group is given by the formula $\binom{m}{2} = \frac{m(m-1)}{2}$, which corresponds to the number of planes containing two distinct coordinate axes in m-dimensional Euclidean space.

The general formulation for a hyperellipsoid based on a Gaussian function in high dimension could be written as:
 \begin{equation} 
	\begin{split}
		G_{(\mu,  \Sigma)}(p) &= \sigma(\alpha)e^{-\frac{1}{2}(p-\mu)^T\Sigma^{-1}(p-\mu)} \\
		\Sigma &= RSS^TR^T \\ 
		R &= R^m \\
		S &= \begin{bmatrix}
			s_1 & 0 & \dots & 0 \\
			0 & s_2 & \dots & 0 \\
            \vdots & \vdots & \ddots & \vdots \\
			0 & 0  & \dots & s_m \\
		\end{bmatrix}
	\end{split}
\end{equation} 
which $R$ is computed using the Equation \ref{eq:r_m} and $S$ is an positive diagonal matrix. 

%%%%%%%%%%%%%%%%%%%%%%%%%%%%%%%%%%%%%%%%%%%%%%%%%%%%%%%%%%%%%%%%%%%%%%%%%%%%%%%
%%%%%%%%%%%%%%%%%%%%%%%%%%%%%%%%%%%%%%%%%%%%%%%%%%%%%%%%%%%%%%%%%%%%%%%%%%%%%%%

\end{document}